\documentclass[10pt,twocolumn,letterpaper]{article}

\usepackage{cvpr}      

\usepackage{graphicx}
\usepackage{amsmath}
\usepackage{amssymb}
\usepackage{booktabs}

\usepackage[T1]{fontenc}
\usepackage{times}
\usepackage{soul}
\usepackage{url}
\usepackage{color}
\usepackage[hidelinks,colorlinks=black, citecolor=black]{hyperref}
\usepackage{graphicx}
\usepackage{amsmath}
\usepackage{booktabs}
\usepackage{algorithm}
\usepackage{algorithmic}

\usepackage{makecell}
\usepackage{epsfig}

\usepackage{bbm}
\usepackage{amsfonts}
\usepackage{mathrsfs}
\usepackage{amsthm,amsmath,amssymb}
\usepackage{multirow}
\usepackage{array}
\usepackage[table,xcdraw]{xcolor}
\usepackage{float}
\usepackage{pifont}
\usepackage{ulem}

\usepackage[capitalize]{cleveref}
\crefname{section}{Sec.}{Secs.}
\Crefname{section}{Section}{Sections}
\Crefname{table}{Table}{Tables}
\crefname{table}{Tab.}{Tabs.}

\begin{document}

\title{Mask Reference Image Quality Assessment}

\author{Pengxiang Xiao\\
\and
Anlong Ming\\
\and
Shuai He\\
\and
Limin Liu\\
\and
{\tt\small Beijing University Posts and Telecommunications 100876,Beijing}
}
\maketitle

\begin{abstract}

Understanding semantic information is an essential step in knowing what is being learned in both full-reference (FR) and no-reference (NR) image quality assessment (IQA) methods. However, especially for many severely distorted images, even if there is an undistorted image as a reference (FR-IQA), it is difficult to perceive the lost semantic and texture information of distorted images \textbf{directly}. In this paper, we propose a \textbf{M}ask \textbf{R}eference IQA (\textbf{MR-IQA}) method that masks specific patches of a distorted image and supplements missing patches with the reference image patches. In this way, our model only needs to input the reconstructed image for quality assessment. First, we design a mask generator to select the best candidate patches from reference images and supplement the lost semantic information in distorted images, thus providing more reference for quality assessment; in addition, the different masked patches imply different data augmentations, which favors model training and reduces overfitting. Second, we provide a \textbf{M}ask \textbf{R}eference \textbf{Net}work (\textbf{MRNet}): the dedicated modules can prevent disturbances due to masked patches and help eliminate the patch discontinuity in the reconstructed image. Our method achieves state-of-the-art performances on the benchmark KADID-10k, LIVE and CSIQ datasets and has better generalization performance across datasets.

\end{abstract}


\section{Introduction}
\label{sec:intro}
\begin{figure}[t]
  \centering
   \includegraphics[width=1.0\linewidth]{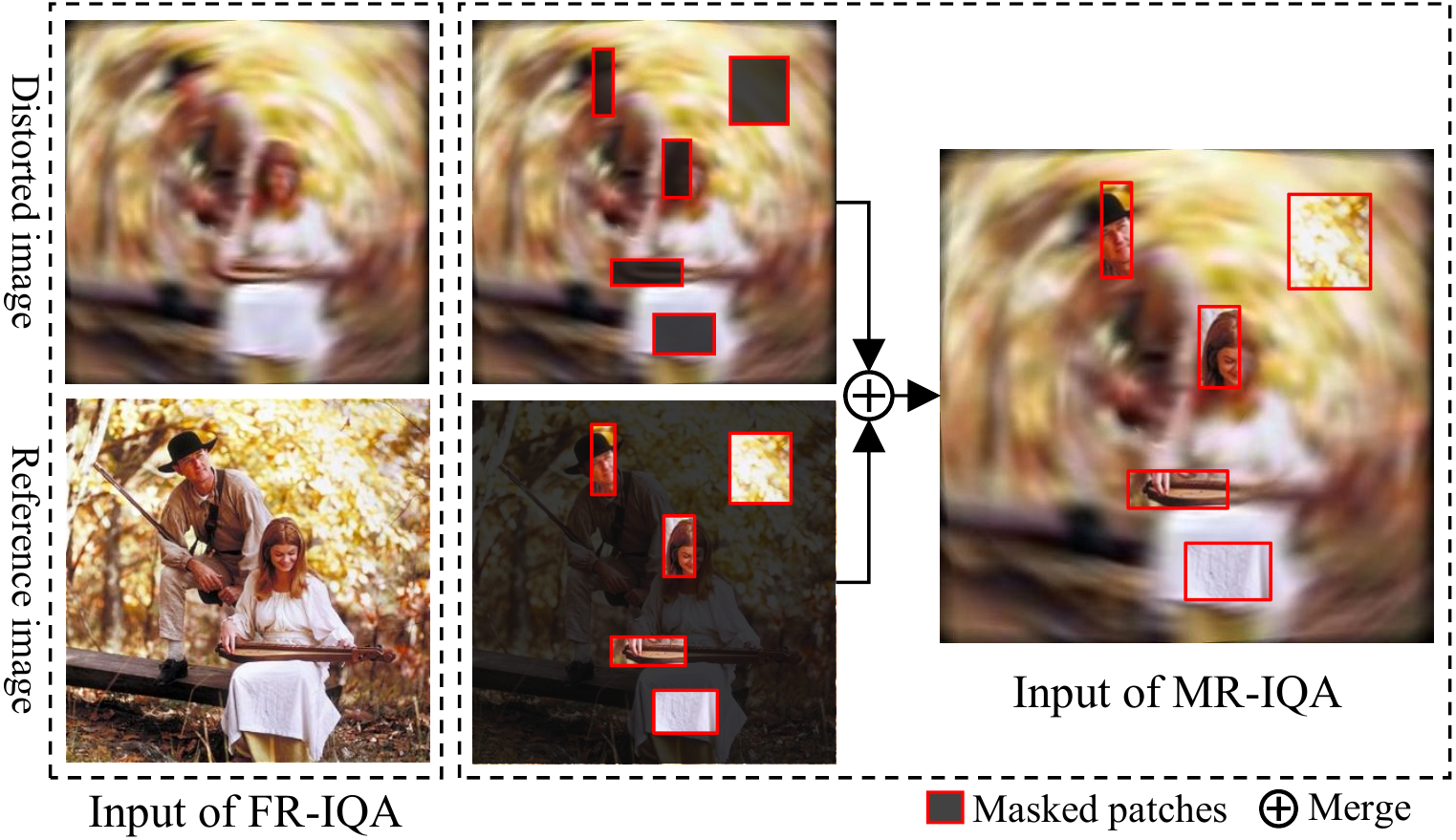}

   \caption{A toy example to demonstrate the input difference between classical FR-IQA and our Mask Reference IQA: the former needs to input a distorted image and the corresponding reference image, and the latter requires only a masked image.}
   \label{fig:introduction}
\end{figure}

Digital images are subject to a wide variety of distortions during acquisition, processing, compression, storage, transmission and reproduction, and each of these processes may result in a degradation in visual quality \cite{mittal2012no} .
The ability to assess and improve the quality of images has become highly desirable in many image processing and computer vision applications. 
However, asking human observers to assess the visual quality requires expensive subjective experiments.
Hence, it is urgent to develop effective methods to automatically evaluate the perceptual quality that are consistent with human subjective perception, and this is known as image quality assessment (IQA).

Based on how much information is available from the reference image, IQA methods can be divided into three categories: full-reference IQA (FR-IQA), reduced-reference IQA (RR-IQA) and no-reference IQA (NR-IQA).
Among them, the FR-IQA methods make full use of the difference between the reference image and the original image for comparison and can better judge the distortion of the image. 
Existing FR-IQA methods usually extract the features from distorted and reference images separately, but these approaches usually fail to directly recover the missing semantic content and texture in distorted images, which is regarded as crucial information for IQA tasks \cite{kim2017deep,hosu2020koniq}.
Moreover, data-driven FR-IQA methods are more complex and have higher computational consumption compared to data-driven NR-IQA methods, and they have to use pairwise labeled data with the mean opinion score (MOS) for training. As the construction process is extremely labor-intensive and costly, available IQA datasets with reference images are too small to be effective for reducing overfitting and training a robust model.

In this paper, we incorporate some patches of the reference image into the distorted images for quality assessment. 
Compared with the previous full reference method, this idea brings two benefits:
(1) The introduced reference image patches can be used to \textbf{directly} recover the necessary but lost semantic and texture information from distorted images to better evaluate image quality (Fig. \ref{fig:introduction}).
(2) The incompleteness of the reference image reduces the dependence of the model on the complete reference image, and the randomness of the image block enables data enhancement and improves the generalization ability of the model.

However, two key challenges remain in designing such a method: 
(1) Which patch of the reference image should be chosen?
It is necessary to ensure that the selected reference image patch can provide sufficient semantic and reference information for the quality assessment of the distorted image while not losing the distorted information in the distorted image. 
(2) How should the supplementary patches of the reference image be utilized? 
It is necessary to ensure that the model can prevent the reference image patches from interfering with the image quality, while extracting necessary semantic and detailed information to help the quality assessment of distorted images.


Driven by this analysis, we present a simple and efficient \textbf{M}ask \textbf{R}eference method for IQA tasks (MR-IQA).
We design a mask generator to select reference image patches, and incorporate them into heavily distorted regions to obtain semantic information gain while maintaining enough distorted information. 
In this way, the semantic information in the distorted image is recovered and reference information is provided for its quality assessment, meanwhile, the random selection of patches can be regarded as a \textbf{natural} data enhancement strategy. 

Then, we refine the Swin Transformer \cite{liu2021swin}, through a dedicated Feature Mask Module (FMM) to shield the perturbation of masked patches in the shallow features, while its patch-wise operations are inherently suitable for exploiting different information in patches. Furthermore, our MR-IQA incorporates a data augmentation strategy to augment datasets while obtaining fixed-size input. In general, the main contributions can be summarized into three-folds:
\begin{itemize}
\item To address the semantic information loss in distorted images, we propose a MR-IQA method. With the dedicated masking strategy, semantic information is recovered in a more direct manner,
while avoiding overfitting on limited datasets. 
\item We introduce a Mask Reference Network (MRNet) with some novel modules to embody the MR-IQA method into an end-to-end model, while also shielding the perturbation of masked patches and augmenting datasets. 
\item Experimental results demonstrate that the proposed MR-IQA not only significantly reduces the computational complexity of the model, but also achieves SOTA performance on multiple datasets. Validation experiments across datasets also demonstrate the strong generalization ability of our method. Moreover, our method has strong generality and can be independently embedded in existing methods or training processes to solve possible stumbling blocks in IQA tasks. 

\end{itemize} 

\section{Related Work}\label{sec：Related Work}

In this section, we briefly introduce some NR-IQA methods, and then review the traditional methods and recent deep learning methods for FR-IQA.
\subsection{No-Reference IQA}
Generally, NR-IQA methods can be divided into natural scene statistics (NSS) based methods and learning-based methods. 
By modeling scene statistics, the traditional NSS based models are sensitive to the appearance of distortion and can detect and quantify the degradation level \cite{mittal2012no,wu2015blind}. 
In recent years, deep learning-based NR-IQA methods have demonstrated superior prediction performance over traditional methods. 
Kang \textit{et al.} \cite{kang2014a} proposed a shallow CNN model and divided the images into several patches to estimate image quality.
Hallucinated-IQA \cite{lin2018hallucinated} estimated the perceptual quality of images with reference images generated by a generative network.
Bosse \textit{et al.} \cite{bosse2017deep} modified VGG-Net \cite{simonyan2014very} to learn a local weight for each image patch to measure the importance of its local quality, and weighted average patch aggregation was adopted as the pooling method. 
Considering the limited size of existing IQA databases may lead to generalization problem, in \cite{bare2017} and \cite{kim2017b}, FR-IQA methods were used to generate the label of patches, and Pan \textit{et al.} utilized the intermediate similarity maps for auxiliary training \cite{pan2019}. 
Meta-IQA \cite{zhu2020metaiqa} and RankIQA \cite{liu2017b} trained the network with separate distortion types for learning prior knowledge. 

Since the IQA is the human visual perception of high-level semantics \cite{gu2014}, models are often pre-trained on ImageNet \cite{deng2009a} to extract semantic features from images \cite{li2016no,bianco2018use,kim2017deep}.
However, the deep semantic features extracted from global scale only represent global information, local details and texture information are ignored. 
To solve above issues, 
Sun \textit{et al.} \cite{sun2016} proposed an NR-IQA framework to combine the global high-level semantics and local low-level characteristics.
Kim \textit{et al.} propose a Multiple-level Feature-based Image Quality Assessor (MFIQA) which considers multiple levels of features simultaneously \cite{Kim2018}. 

Although the NR-IQA methods get rid of the dependence on reference images and obtain partial semantic information, they still cannot achieve the competitive performance of FR-IQA methods due to the lack of reference information.
\subsection{Full-Reference IQA}
FR-IQA methods compare the distorted image against its pristine-quality reference, which can also be further divided into traditional evaluation metrics and learning-based models. Traditional evaluation metrics correlate image quality with some hand-crafted definitions of perceptual differences between the inputs.
The most widely used are mean squared error (MSE) and peak signal-to-noise ratio (PSNR). MSE is a signal-based metric that represents the cumulative squared error between the distorted and the reference image. PSNR is the most popular pixel-based metric which represents a measure of the peak error.
Image fidelity criterion (IFC) \cite{sheikh2005information} and visual information fidelity (VIF) \cite{sheikh2006image} are natural scene statistics (NSS) based metrics that model natural images in the wavelet domain. Structure similarity (SSIM) \cite{wang2004image} and MS-SSIM \cite{wang2003multiscale} methods considered the human vision system (HVS) and utilized the local structure similarity to evaluate image quality.

As visual perception is a complicated process, it is difficult to simulate the HVS with limited hand-crafted features. 
To solve this problem, learning-based FR-IQA models use deep networks to extract features from training data without expert knowledge.
Amirshahi \textit{et al.} \cite{amirshahi2016image} used the pre-trained model to capture the feature maps of the test and reference images at multiple layers and compare their feature similarity at each layer. 
Zhang \textit{et al.} \cite{Zhang_2018_CVPR} proposed the Learned Perceptual Image Patch Similarity (LPIPS) to obtain the perceptual similarity judgment by calculating the Euclidean distance between the reference and distorted deep feature representations.
Similarly, Ding \textit{et al.} \cite{ding2020image} presented a Deep Image Structure and Texture Similarity metric (DISTS) to measure the similarities between the VGG-based deep features. 
Hammou \textit{et al.} \cite{hammou2021egb} proposed an ensemble of gradient boosting (EGB) metrics based on selected feature similarity and ensemble learning. 
Peng \textit{et al.} designed a Multi-Metric Fusion Network (MMFN) \cite{Peng_2021_CVPR} for aggregating the quality scores predicted by diverse metrics to generate more accurate results.

In recent years, some researchers have tried to apply Transformer in IQA tasks to extract more powerful features.
IQT \cite{cheon2021perceptual} extracted the feature from CNNs and then fed the feature maps into the transformer for the quality prediction task. 
SwinIQA \cite{liu2022swiniqa} used the Swin Transformer \cite{liu2021swin} to extract features and measured the perceptual quality of compressed images in a learned Swin distance space.

However, the above FR-IQA methods all extract features from the reference image and the distorted image separately, which will cause two problems:
On the one hand, the model cannot directly obtain the lost semantic and texture information when extracting features from distorted images (Fig. \ref{fig:introduction}), \textit{e.g.}, 
the neglect of the high-level semantic information may result in predicting a clear blue sky as a bad quality, which is inconsistent with human perception \cite{li2017}.
On the other hand, compared with the no-reference IQA method, using the model to process a pair of images will result in costing twice the amount of computation. In this paper, we present a simple general MR-IQA method to solve above problems.

\begin{figure*}[t]
  \centering   
   \includegraphics[width=1.0\linewidth]{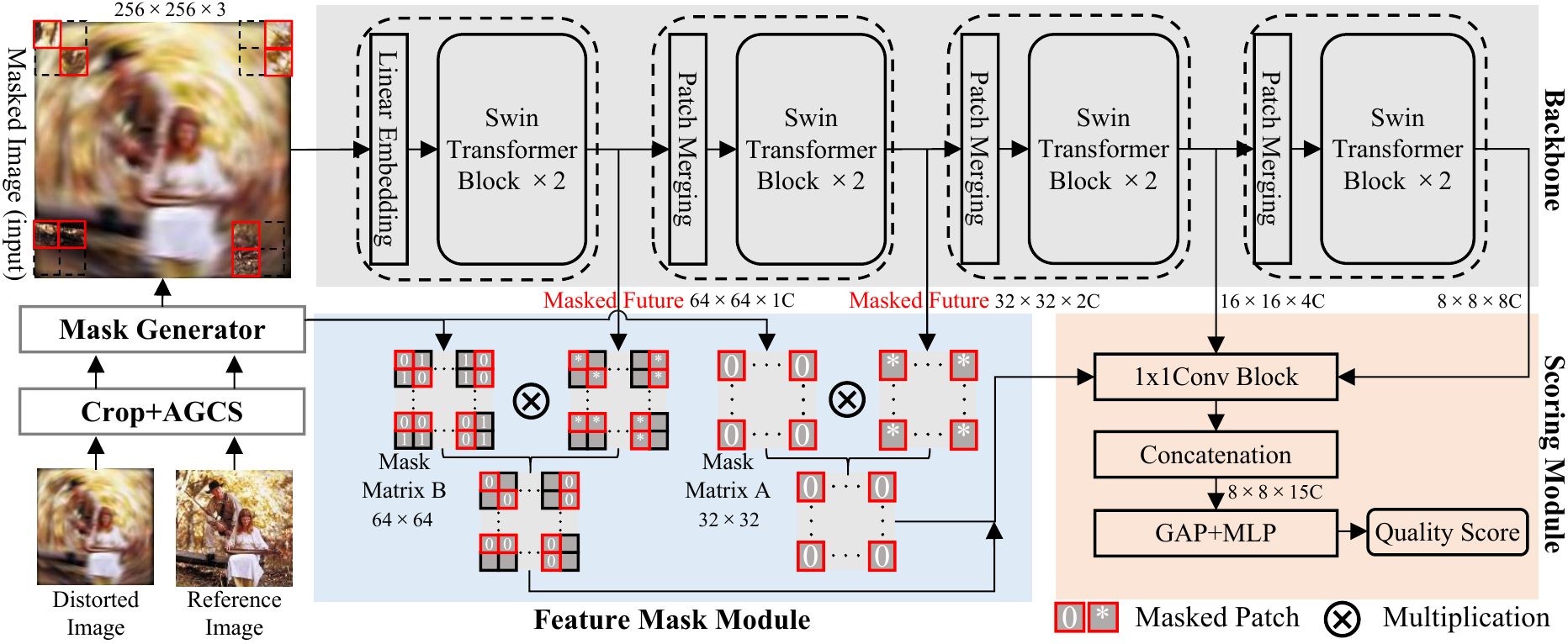}

   \caption{Illustration of the proposed MR-IQA. Given the distorted image and reference image, we first preprocess the two images via the Adaptive Grid Cropping and Sampling (AGCS) module to obtain fixed-size images, and feed them into the Mask Generator (MG) module to obtain the masked image. Then, the masked image is sent to the backbone to obtain multilayer features, specifically, the shallow features shield the effect of masked features through Feature Mask Module (FMM). Finally, the processed features are sent to scoring module to complete the quality scoring.}
   \label{fig:structure}
\end{figure*}

\section{Method}


In this section, we describe the full pipeline of our proposed Mask-Reference IQA method. As shown in Fig. \ref{fig:structure}, reference images and distorted images are first cropped and sampled into fragments via the \textbf{A}daptive \textbf{G}rid \textbf{C}ropping and \textbf{S}ampling (\textbf{AGCS}) module to obtain fixed-size images. Then, the preprocessed images are fed into the \textbf{M}ask \textbf{G}enerator (\textbf{MG}) module to generate the masked images as input. Finally, the masked images are sent to the \textbf{M}ask \textbf{R}eference \textbf{N}etwork (\textbf{MRNet}) to get final predictions of image quality. We describe details as follows.




\subsection{Adaptive Grid Cropping and Sampling}
In this paper, we do not resize or use a smaller crop size to meet the model input size like other FR-IQA methods, since resizing corrupts local textures and cropping with small size causes mismatched global quality with local regions. 
Instead, we directly use a larger size to crop patches from images and design the AGCS module (Fig. \ref{fig:Grid Sample}) to well preserve the original image quality and get output with a fixed size. 
The core design is dividing the patch obtained by random crop into uniform grids with the same size, and then obtaining fixed-size outputs by cropping and sampling grids adaptively. 

Given the distorted images $I_{dst}$ and the corresponding reference images $I_{ref}$, we first use a random crop operation to get $P_{dst}$ and $P_{ref}$
with the same position and size $W_p \times H_p$. Since $P_{dst}$ and $P_{ref}$ follow the same processing flow, we use $P$ to represent them.
It should be noted that the crop operation uses a larger random aspect ratio size, where the length and width are both integer multiples of 64 to facilitate subsequent processing.
In this way, larger patches can capture a wider range of semantic information, and direct processing of the raw resolution image results in no loss of the local textures, which are vital in IQA.



Then, we divided $P$ into $G_w \times G_h$ grids with the same sizes, denoted as $G= \left \{ g^{0,0}, ..., g^{i,j}, ..., g^{G_w-1,G_h-1}\right \}$, it can be formalized as follows:  

\begin{equation}
    W_g= \frac{W_p}{G_w} \ , \ H_g= \frac{H_p}{G_h},
\end{equation}
\begin{gather}
    g^{i,j}= P\left[ \right. { {i \times W_{g}}} :{{\left ( i+1 \right ) \times W_g}} \ {,}\    
      { {j \times H_{g}}} :{{\left ( j+1 \right ) \times H_g}} \left.\right ], \nonumber\\
     {0 \leq i < G_w \, {,} \, 0 \leq j < G_h} ,
\end{gather}
where $W_{g}$ and $H_{g}$ denote the width and height of each grid, $g^{i,j}$ denotes the area includes by the $i$-th row and $j$-th column grid in $P$. In this step, we obtain grid as the minimum operating unit for cropping and sampling adaptively. 

Finally, we randomly crop each $g^{i,j}$ at the same position, to avoid disrupting the original semantic information due to the different distances between grids. The cropping process in a grid is formalized as follows:


\begin{equation}
    W_{g^{\prime}}= \frac{W_{input}}{G_w} \ , \  H_{g^{\prime}}= \frac{H_{input}}{G_h},
\end{equation}
\begin{gather}
     g^{\prime \, i,j}=g^{i,j} \left[\right. {m_{ran}}:{m_{ran}+W_{g^{\prime}}}, 
     \left. {n_{ran}}:{n_{ran}+H_{g^{\prime}}} \right] , \nonumber\\
     {0 \leq m_{ran} < W_{g}-W_{g^{\prime}} \, {,} \, 0 \leq n_{ran} < H_g-H_{g^{\prime}}} ,
\end{gather}
where $m_{ran}$ and $n_{ran}$ are random positions to crop in each grid.
In this way, girds of patches with different sizes and regions in the image are randomly selected, which prevents overfitting caused by the small dataset.
$W_{g^{\prime}}$ and $H_{g^{\prime}}$ denote the width and height of the $g^{\prime}$, which are calculated based on the model desired input size $W_{input} \times H_{input}$ and number of grids $G_{w} \times G_{h}$. 
In our method, the number of grids $G_{w} \times G_{h}$ is set to $64 \times 64$, the size of of the $g^{\prime}$ will be adaptively changed based on the patch size, so as to ensure that the size of the $G^{\prime}= \left \{ g^{\prime \, 0,0},..., g^{\prime \, i,j},..., g^{\prime \, G_w-1,G_h-1}\right \}$ is fixed.


After the above pipeline, $G^{\prime}_{dst}$ and $G^{\prime}_{ref}$ with fixed-size are obtained from $P_{dst},P_{ref}$
for further mask processing in Mask Generator (MG) module.

\subsection{Mask Generator Module}
To generate a mask image for quality assessment, we mask severely distorted patches and supplement patches of reference images at the same position (Fig. \ref{fig:MG module}). First, we use Mean Absolute Error (\textbf{MAE}) to estimate the difference $Diff$ between the $G^{\prime}_{dst}$ and $G^{\prime}_{ref}$ at each location:
\begin{gather}
    Diff^{i,j}=\textbf{MAE}\left (  g_{dst}^{\prime \, i,j} , g_{ref}^{\prime \, i,j} \right ){,}  \nonumber\\
    {0 \leq i < G_w \, {,} \, 0 \leq j < G_h} ,
\end{gather}
where $Diff^{i,j}$ represents the distortion degree of $G^{\prime}_{dst}$ at the $(i,j)$ position, the larger the value, the greater the distortion of the $G^{\prime}_{dst}$ at this position. We select masked patches based on the following principles: select regions with higher perceptual differences between the reference and the distorted images, since patches that are similar include little information about image quality differences, and can only supplement limited semantic and texture information. To do this, we calculate the median $mid$ in the $Diff^{i,j}$, then in the $Diff^{i,j}$, values greater than the median are set to 1; otherwise, they are set to 0. $Diff^{i,j}$ is redefined as $Mask_a$, which only includes 0 or 1 values. The process is formalized as follows:
\begin{figure}[t]
  \centering
   \includegraphics[width=0.8\linewidth]{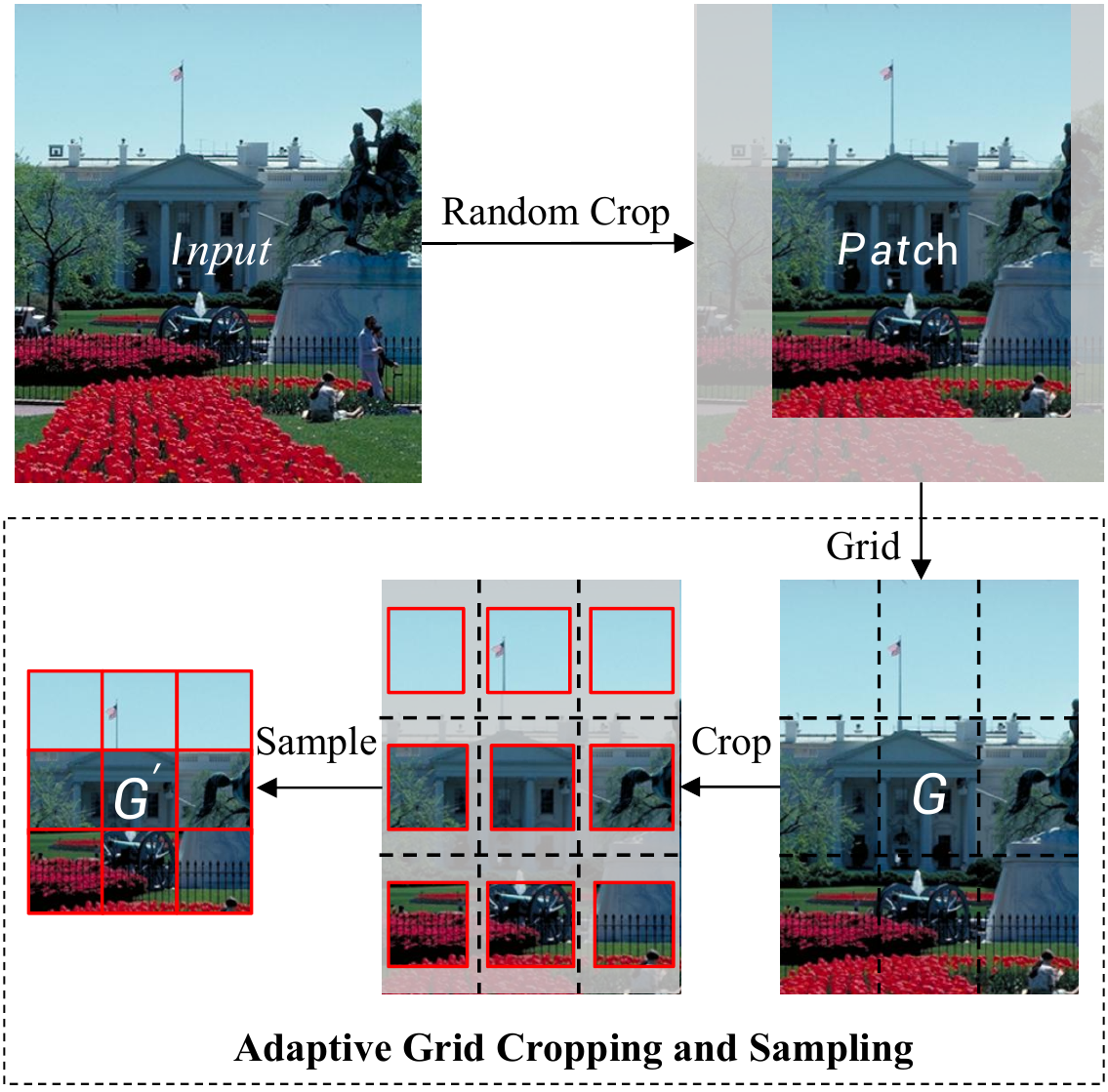}
   \caption{To obtain fixed-size output while augmenting dataset, in our Adaptive Grid Cropping and Sampling (AGCS) module, the input will be cropped into a patch of sufficient size, and then some grids generate in the patch and sampled randomly by the adaptive sampling strategy to fix output size.}
   \label{fig:Grid Sample}
\end{figure}
\begin{equation}
    Mask_a^{i,j}=\left\{\begin{matrix}
     &1, &if \quad {Diff^{i,j}>mid}  \\ 
     &0,  &else 
    \end{matrix}\right. .
\end{equation}
The position where the value is 1 in $Mask_a$ indicates the semantic and detail information loss is more serious than other positions in the distorted image.

Second, although it is necessary for quality assessment to supplement the lost semantic information in mask-selected locations in the distorted image, these regions also contain distortion information, which cannot be lost.
In order to solve this contradiction, we randomly select the positions where $Mask_a^{i,j}=1$ to ensure that half of each selected mask is reserved for judgment distortion, and the remaining half is replaced by reference patches.
In this way, each time an image is processed, a new and unseen random set of patches of the reference image is masked into the distorted image.
This provides the hidden benefit of data augmentation, which aids in model training and reduces overfitting.

Specifically, we randomly generate a binary feature map $Mask_{ran}$, in which the proportion of 0 and 1 values is same.
$Mask_{ran}$ is twice the size of $Mask_a$, so $Mask_a$ need to be upsampled to the same size as $Mask_{ran}$ by copying.
After that, we compute the intersection of these masks to obtain $Mask_{b}$:
\begin{equation}
    Mask_{b}^{i,j}=Mask_{a}^{i,j} \bigcap Mask_{ran}^{i,j} ,
\end{equation}
according to the probability calculation, nearly 25\% of the values in $Mask_{b}$ are 1, and the rest are 0.

Finally, $G_{dst}^{\prime}$ and $G_{ref}^{\prime}$ are merged to get $G_{mask}$ based on $Mask_{b}$, each grid $g_{mask}^{i,j}$ of $G_{mask}$ is obtained as follows:
\begin{equation}
    g_{mask}^{i,j}=\left\{\begin{matrix}
     &g_{ref}^{\prime \, i,j}, &if \quad {Mask_b^{i,j}=1}  \\ 
     &g_{dst}^{\prime \, i,j}, &else 
    \end{matrix}\right. .
\end{equation}

Existing works \cite{gu2014} have shown that the global semantic information affects the quality predictions. 
Therefore, we splice each grid in $G_{mask}$ into their original positions to generate masked images $I_{mask}$, which are fed into MRNet for final quality assessment.
\begin{figure}[t]
  \centering
   \includegraphics[width=1.0\linewidth]{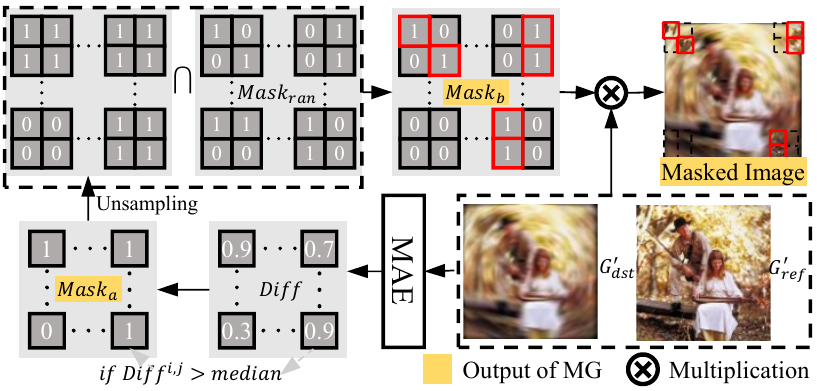}

   \caption{The pipeline of the Mask Generator: firstly, we calculate the difference between the reference and distorted image (preprocessed by the AGCS module) based on mean absolute error, then filter out the areas with larger difference (> median) and mark them as "1", and finally replace these areas in the distorted image with the corresponding areas of the reference image, to form the masked image.}
   \label{fig:MG module}
\end{figure}




\subsection{Mask Reference Network}

As shown in Fig. \ref{fig:structure}, our proposed MRNet consists of a backbone to extract multilayer features, a \textbf{F}eature \textbf{M}ask \textbf{M}odule (\textbf{FMM}) to process features and a Scoring Module to predict the quality score after fusing the features.
When the AGCS module merges the reference patches into the distorted images to restore the semantic information and reference information, these patches also cause a spatial discontinuity of the masked images. 
The CNN-based models use convolutional layers to extract features from patches in different locations, 
which may make it difficult for the model to distinguish the artificial discontinuity between patches, even confusing it with quality degradation.

As a hierarchical structure based on patch-wise operations, the Swin Transformer \cite{liu2021swin} is more suitable for processing patch-based input than their CNN counterparts. 
For the task of this paper, on the one hand, Transformer (pre-trained on ImageNet \cite{deng2009a}) have a global receptive field that can help restore the lost semantic information from limited reference patches. 
On the other hand, the non-local self-attention structure included in the Transformer can compare the difference between the local reference block and other regions, thus better evaluating the specific distortion degree of the distorted region. 
Therefore, we used the Swin Transformer pre-trained on the ImageNet as the backbone, and extracted multi-scale features $F_k$ (k$\in$ {1, 2, 3, 4}), which are obtained from the four stages to capture both global semantic information and local detail information (Fig. \ref{fig:structure}).
To further avoid the interference of the reference image patches in the masked image to the quality assessment of the distorted image, 
we design the Feature Mask Module to erase the information from the reference image in the output of stage-1 and stage-2. Specifically, we process $F_1 \in \mathbb{R}^{64 \times 64 \times C}$ and $F_2 \in \mathbb{R}^{32 \times 32 \times C}$ based on $Mask_a \in \mathbb{R}^{32 \times 32}$ and $Mask_b \in \mathbb{R}^{64 \times 64}$ as follows:
\begin{equation}
    F_1^{i,j}=\neg Mask_b^{i,j} \times F_1^{i,j}({0 \leq {i,j} < 64}),
\end{equation}
\begin{equation}
    F_2^{i,j}=\neg Mask_a^{i,j} \times F_2^{i,j}({0 \leq {i,j} < 32}).
\end{equation}
The $F_k$ (k$\in$ {1, 2}) denote the feature vector at the position $(i, j)$.
The ${Mask_{a}^{i,j}}$ and ${Mask_{b}^{i,j}}$ are outputted from MG, after bit inversion processing they denote whether the distorted image patches at $(i,j)$ is masked. Considering that the deep features outputted by the stage-3 and stage-4 in Swin Transformer contain abstract semantic information and distortion features, no processing is done on these features.


Finally, we use the Scoring Module to map the multilayer features to quality scores.
As shown in Figure 2, 
Adaptive average pooling (GAP) is used to align sizes of four multiscale features firstly, 
then we use $1 \times 1$ convolution block to boost the channel interaction among the extracted features, 
and concatenate the features along the channel dimension. 
After these features are globally average pooled into vectors, with a sigmoid function as the activation function, MLP which consists of two fully connected layers is employed to regress the vectors to the quality score.
\section{Experiments}\label{sec:Experiment}
In this section, we first introduce experimental settings including IQA datasets, evaluation criteria and implementation details of our method. Then, we compare it with state-of-the-art FR and NR IQA methods on four benchmark datasets, and evaluate the generalization ability of our method. 
In addition, we also compare the impact of different mask ratios on performance. 
Finally, we conduct ablation studies to analyze the proposed method. 


\subsection{Experimental Settings}
\begin{table}[]
\setlength{\tabcolsep}{0.6mm}{
\small
\begin{center}
\begin{tabular}{ccccccc}
\Xhline{1.2pt}
 \hline \specialrule{0em}{1pt}{1pt}
Dataset   & Ref. & Dis. & Dis.type & Rat. & Rat.type & Score    \\ \hline \specialrule{0em}{1pt}{1pt}
\cellcolor[HTML]{EFEFEF}LIVE\cite{sheikh2003image}      & \cellcolor[HTML]{EFEFEF}29     & \cellcolor[HTML]{EFEFEF}779    & \cellcolor[HTML]{EFEFEF}5          & \cellcolor[HTML]{EFEFEF}25k      & \cellcolor[HTML]{EFEFEF}DMOS        & \cellcolor[HTML]{EFEFEF}{[}0,100{]}    \\ \specialrule{0em}{1pt}{1pt}
CSIQ\cite{larson2010most}      & 30     & 866    & 6          & 5k       & DMOS        & {[}0,1{]}      \\ \specialrule{0em}{1pt}{1pt}
\cellcolor[HTML]{EFEFEF}TID2013\cite{ponomarenko2015image}   & \cellcolor[HTML]{EFEFEF}25     & \cellcolor[HTML]{EFEFEF}3000   & \cellcolor[HTML]{EFEFEF}24         & \cellcolor[HTML]{EFEFEF}524k     & \cellcolor[HTML]{EFEFEF}MOS         & \cellcolor[HTML]{EFEFEF}{[}0,9{]}      \\ \specialrule{0em}{1pt}{1pt}
KADID-10k\cite{lin2019kadid} & 81     & 10125  & 25         & 30.4k    & MOS         & {[}1,5{]}      \\ \specialrule{0em}{1pt}{1pt}
\Xhline{1.2pt}
\hline 
\end{tabular}
\caption{Summary of four IQA databases: LIVE \cite{sheikh2003image}, CSIQ\cite{larson2010most}, TID2013\cite{ponomarenko2015image} and KADID-10K\cite{lin2019kadid}. }
\label{table_dataset}
\end{center}}

\end{table}

\begin{table}[]
\scriptsize
\begin{center}
\setlength{\tabcolsep}{0.65mm}{
\begin{tabular}{c|l|cc|cc|cc|cc}
\Xhline{1.2pt}
\hline
\specialrule{0em}{1pt}{1pt}
                      & \multicolumn{1}{c|}{}                          & \multicolumn{2}{c|}{LIVE}                                     & \multicolumn{2}{c|}{CSIQ}                                     & \multicolumn{2}{c|}{TID2013}                                  & \multicolumn{2}{c}{KADID-10k}                                 \\\specialrule{0em}{1pt}{1pt}
\multirow{-2}{*}{}    & \multicolumn{1}{c|}{\multirow{-2}{*}{Methods}} & SRCC                         & PLCC                        & SRCC                         & PLCC                        & SRCC                         & PLCC                        & SRCC                         & PLCC                        \\  \hline \specialrule{0em}{1pt}{1pt}
                      & \cellcolor[HTML]{EFEFEF}BRISQUE\cite{mittal2012no}                & \cellcolor[HTML]{EFEFEF}0.939 & \cellcolor[HTML]{EFEFEF}0.935 & \cellcolor[HTML]{EFEFEF}0.746 & \cellcolor[HTML]{EFEFEF}0.829 & \cellcolor[HTML]{EFEFEF}0.604 & \cellcolor[HTML]{EFEFEF}0.694 & \cellcolor[HTML]{EFEFEF}0.528     & \cellcolor[HTML]{EFEFEF}0.567     \\ \specialrule{0em}{1pt}{1pt}
                      & FRIQUEE\cite{ghadiyaram2017perceptual}                                        & 0.940                         & 0.944                         & 0.835                         & 0.874                         & 0.680                         & 0.753                         & -                             & -                             \\ \specialrule{0em}{1pt}{1pt}
                      & \cellcolor[HTML]{EFEFEF}CORNIA\cite{ye2012unsupervised}                 & \cellcolor[HTML]{EFEFEF}0.947 & \cellcolor[HTML]{EFEFEF}0.950 & \cellcolor[HTML]{EFEFEF}0.678 & \cellcolor[HTML]{EFEFEF}0.776 & \cellcolor[HTML]{EFEFEF}0.678 & \cellcolor[HTML]{EFEFEF}0.768 & \cellcolor[HTML]{EFEFEF}0.541 & \cellcolor[HTML]{EFEFEF}0.580 \\ \specialrule{0em}{1pt}{1pt}
                      & M3\cite{xue2014blind}                                             & 0.951                         & 0.950                         & 0.795                         & 0.839                         & 0.689                         & 0.771                         & -                             & -                             \\ \specialrule{0em}{1pt}{1pt}
                      & \cellcolor[HTML]{EFEFEF}HOSA\cite{xu2016blind}                   & \cellcolor[HTML]{EFEFEF}0.946 & \cellcolor[HTML]{EFEFEF}0.947 & \cellcolor[HTML]{EFEFEF}0.741 & \cellcolor[HTML]{EFEFEF}0.823 & \cellcolor[HTML]{EFEFEF}0.735 & \cellcolor[HTML]{EFEFEF}0.815 & \cellcolor[HTML]{EFEFEF}0.609 & \cellcolor[HTML]{EFEFEF}0.653 \\ \specialrule{0em}{1pt}{1pt}
                      & BIECON\cite{kim2016fully}                 & 0.961 & 0.962 & 0.815 & 0.823 & 0.717 & 0.762 & 0.623     & 0.648     \\ \specialrule{0em}{1pt}{1pt}
                      & \cellcolor[HTML]{EFEFEF}WaDIQaM\cite{bosse2017deep}                & \cellcolor[HTML]{EFEFEF}0.954 & \cellcolor[HTML]{EFEFEF}0.963 & \cellcolor[HTML]{EFEFEF}0.844     & \cellcolor[HTML]{EFEFEF}0.852     & \cellcolor[HTML]{EFEFEF}0.761 & \cellcolor[HTML]{EFEFEF}0.787 & \cellcolor[HTML]{EFEFEF}0.739     & \cellcolor[HTML]{EFEFEF}0.752     \\ \specialrule{0em}{1pt}{1pt}
                      & ResNet-ft\cite{kim2017deep}                                      & 0.950                         & 0.954                         & 0.876                         & 0.905                         & 0.712                         & 0.756                         & -                             & -                             \\ \specialrule{0em}{1pt}{1pt}
                      & \cellcolor[HTML]{EFEFEF}IW-CNN\cite{kim2017deep}                 & \cellcolor[HTML]{EFEFEF}0.963 & \cellcolor[HTML]{EFEFEF}0.964 & \cellcolor[HTML]{EFEFEF}0.812 & \cellcolor[HTML]{EFEFEF}0.791 & \cellcolor[HTML]{EFEFEF}0.800 & \cellcolor[HTML]{EFEFEF}0.802 & \cellcolor[HTML]{EFEFEF}-     & \cellcolor[HTML]{EFEFEF}-     \\ \specialrule{0em}{1pt}{1pt}
                      & CaHDC\cite{wu2020end}                  & 0.965 & 0.964 & 0.903 & 0.914 & 0.862 & 0.878 & -     & -     \\ \specialrule{0em}{1pt}{1pt}
\multirow{-13}{*}{\rotatebox{90}{NR}} & \cellcolor[HTML]{EFEFEF}HyperIQA\cite{su2020blindly}                                       & \cellcolor[HTML]{EFEFEF}0.962                         & \cellcolor[HTML]{EFEFEF}0.966                         & \cellcolor[HTML]{EFEFEF}0.923                         & \cellcolor[HTML]{EFEFEF}0.942                         & \cellcolor[HTML]{EFEFEF}0.729                         & \cellcolor[HTML]{EFEFEF}0.775                         & \cellcolor[HTML]{EFEFEF}0.852                             & \cellcolor[HTML]{EFEFEF}0.845                             \\  \hline \specialrule{0em}{1pt}{1pt}
                      & PSNR                   & 0.873 & 0.865 & 0.810 & 0.819 & 0.687 & 0.677 & 0.676 & 0.675 \\ \specialrule{0em}{1pt}{1pt}
                      & \cellcolor[HTML]{EFEFEF}SSIM\cite{wang2004image}                                           & \cellcolor[HTML]{EFEFEF}0.948                         & \cellcolor[HTML]{EFEFEF}0.937                         & \cellcolor[HTML]{EFEFEF}0.865                         & \cellcolor[HTML]{EFEFEF}0.852                         & \cellcolor[HTML]{EFEFEF}0.727                         & \cellcolor[HTML]{EFEFEF}0.777                         & \cellcolor[HTML]{EFEFEF}0.724                         & \cellcolor[HTML]{EFEFEF}0.717                         \\ \specialrule{0em}{1pt}{1pt}
                      & MS-SSIM\cite{wang2003multiscale}                & 0.951 & 0.940 & 0.906 & 0.889 & 0.786 & 0.830 & 0.826 & 0.820 \\ \specialrule{0em}{1pt}{1pt}
                      & \cellcolor[HTML]{EFEFEF}VSI\cite{zhang2014vsi}                                            & \cellcolor[HTML]{EFEFEF}0.952                         & \cellcolor[HTML]{EFEFEF}0.948                         & \cellcolor[HTML]{EFEFEF}0.942                         & \cellcolor[HTML]{EFEFEF}0.928                         & \cellcolor[HTML]{EFEFEF}0.897                         & \cellcolor[HTML]{EFEFEF}0.900                         & \cellcolor[HTML]{EFEFEF}0.879                         & \cellcolor[HTML]{EFEFEF}0.877                         \\ \specialrule{0em}{1pt}{1pt}
                      & FSIMc\cite{zhang2011fsim}                  & 0.965 & 0.961 & 0.931 & 0.919 & 0.851 & 0.877 & 0.854 & 0.850 \\ \specialrule{0em}{1pt}{1pt}
                      & \cellcolor[HTML]{EFEFEF}MAD\cite{larson2010most}                                            & \cellcolor[HTML]{EFEFEF}0.967                         & \cellcolor[HTML]{EFEFEF}0.968                         & \cellcolor[HTML]{EFEFEF}0.947                         & \cellcolor[HTML]{EFEFEF}0.950                         & \cellcolor[HTML]{EFEFEF}0.781                         & \cellcolor[HTML]{EFEFEF}0.827                         & \cellcolor[HTML]{EFEFEF}0.799                         & \cellcolor[HTML]{EFEFEF}0.799                         \\ \specialrule{0em}{1pt}{1pt}
                      & VIF\cite{sheikh2006image}                    & 0.964 & 0.960 & 0.911 & 0.913 & 0.677 & 0.771 & 0.679 & 0.687 \\ \specialrule{0em}{1pt}{1pt}
                      & \cellcolor[HTML]{EFEFEF}WaDIQaM\cite{bosse2017deep}                                        & \cellcolor[HTML]{EFEFEF}0.970                         & \cellcolor[HTML]{EFEFEF}0.980                         & \cellcolor[HTML]{EFEFEF}0.901                             & \cellcolor[HTML]{EFEFEF}0.909                             & \cellcolor[HTML]{EFEFEF}\textbf{0.940}                         & \cellcolor[HTML]{EFEFEF}\textbf{0.946}                         & \cellcolor[HTML]{EFEFEF}0.896                             & \cellcolor[HTML]{EFEFEF}0.889                             \\ \specialrule{0em}{1pt}{1pt}
                      & DISTS\cite{ding2020image}                  & 0.955 & 0.955 & 0.946 & 0.946 & 0.830 & 0.855 & 0.887 & 0.886 \\ \specialrule{0em}{1pt}{1pt}
                      & \cellcolor[HTML]{EFEFEF}PieAPP\cite{prashnani2018pieapp}                                         & \cellcolor[HTML]{EFEFEF}0.918                         & \cellcolor[HTML]{EFEFEF}0.909                         & \cellcolor[HTML]{EFEFEF}0.890                         & \cellcolor[HTML]{EFEFEF}0.873                         & \cellcolor[HTML]{EFEFEF}0.876                         & \cellcolor[HTML]{EFEFEF}0.859                         & \cellcolor[HTML]{EFEFEF}0.836                         & \cellcolor[HTML]{EFEFEF}0.836                         \\ \specialrule{0em}{1pt}{1pt}
                      \multirow{-13}{*}{\rotatebox{90}{FR}} &LPIPS\cite{zhang2018unreasonable}                  & 0.932 & 0.934 & 0.903 & 0.927 & 0.670 & 0.749 & 0.843 & 0.839 \\ \specialrule{0em}{1pt}{1pt}
\hline \specialrule{0em}{1pt}{1pt}
                      & \cellcolor[HTML]{EFEFEF}Ours (Random)                                           & \cellcolor[HTML]{EFEFEF}0.976                         & \cellcolor[HTML]{EFEFEF}0.979                         & \cellcolor[HTML]{EFEFEF}0.943                         & \cellcolor[HTML]{EFEFEF}0.949                         & \cellcolor[HTML]{EFEFEF}0.904                         & \cellcolor[HTML]{EFEFEF}0.925                         & \cellcolor[HTML]{EFEFEF}0.949                         & \cellcolor[HTML]{EFEFEF}0.952                         \\
                      \specialrule{0em}{1pt}{1pt}
                      & Ours (Diff)                   & \textbf{0.977} & \textbf{0.980} & \textbf{0.947} & \textbf{0.952} & 0.912 & 0.930 & \textbf{0.952} & \textbf{0.955} \\ \Xhline{1.2pt} \hline \specialrule{0em}{1pt}{1pt}
                    
\end{tabular}
\caption{Performance evaluation on the LIVE \cite{sheikh2003image}, CSIQ\cite{larson2010most}, TID2013\cite{ponomarenko2015image} and KADID-10K\cite{lin2019kadid}, our proposed MR-IQA achieves SOTA performance on three datasets.}
\label{table_performance}
}

\end{center}
\end{table}
\noindent\textbf{Evaluation Datasets.}
The main experiments are conducted on four singly distorted synthetic IQA databases: LIVE \cite{sheikh2003image}, CSIQ \cite{larson2010most}, TID2013 \cite{ponomarenko2015image} and KADID-10K \cite{lin2019kadid},
whose configurations are presented in Table \ref{table_dataset}. 
LIVE, CSIQ and TID2013 are three relatively small-scale IQA datasets, where distorted images only contain traditional distortion types (e.g., noise, downsampling, JPEG compression). 
KADID-10K \cite{lin2019kadid} further incorporates the recovered results of a denoising algorithm into the distorted images, resulting in a medium-sized IQA dataset.
Since the explicit splits of training and testing are not given on these four datasets,
we randomly split 80\% distorted images for training and the rest 20\% ones for testing like previous works do. 
It should be emphasized that our split is based on reference images to avoid content overlapping. 
To reduce the bias caused by a random split, we run the random train-test splitting operation ten times, the comparison results are reported as the average of ten times evaluation experiments.

\noindent\textbf{Evaluation Metrics.}
To evaluate the performance of the proposed method, the Spearman’s Rank Ordered Correlation Coefficient (SRCC) and the Pearson’s Linear Correlation Coefficient (PLCC) are applied between the subjective DMOS/MOS from human opinion and the predicted score. The two criteria both range from 0 to 1 and a higher value indicates better performance. 
\subsection{Implementation Details}
As reported in \cite{dosovitskiy2020image}, Vision Transformer strongly benefits from pre-training on larger datasets prior to transfer to downstream tasks. 
To avoid training Transformer from scratch, 
we use the Swin Transformer (Tiny) which is pre-trained on ImageNet-1k \cite{deng2009a} classification datasets for all our experiments. 
We set initial learning rate to $1e^{-6}$ with a decay rate of 0.5 after every 20 epochs, using Adam optimizer \cite{kingma2014adam} and the mini-batch size is set to 16.
Our experiments are implemented on an NVIDIA GeForce RTX 3090 with PyTorch 1.8.0 and CUDA 11.2 for training and testing.


Thanks to the adaptive sampling of images by the AGCS module, we do not limit the size of the input images through resizing or cropping during training and testing.
However, in order to ensure that the model performance is not affected, under the premise of not exceeding the size of the input image, we designed a variety of possible crop sizes (256, 320, 384, 448, 512) for random selection. 
After cropping, 8 patches from test image are randomly sampled and perform flipping (horizontal/vertical) with a given probability 0.5, their corresponding prediction scores are average pooled to get the final quality score.
We evaluate the training of our model with MAE and MSE loss function, and find that more consistent results are obtained with MAE loss.

\subsection{Comparison with the State-of-the-art Methods}
\begin{figure}[t]
  \centering
    \includegraphics[width=1.0\linewidth]{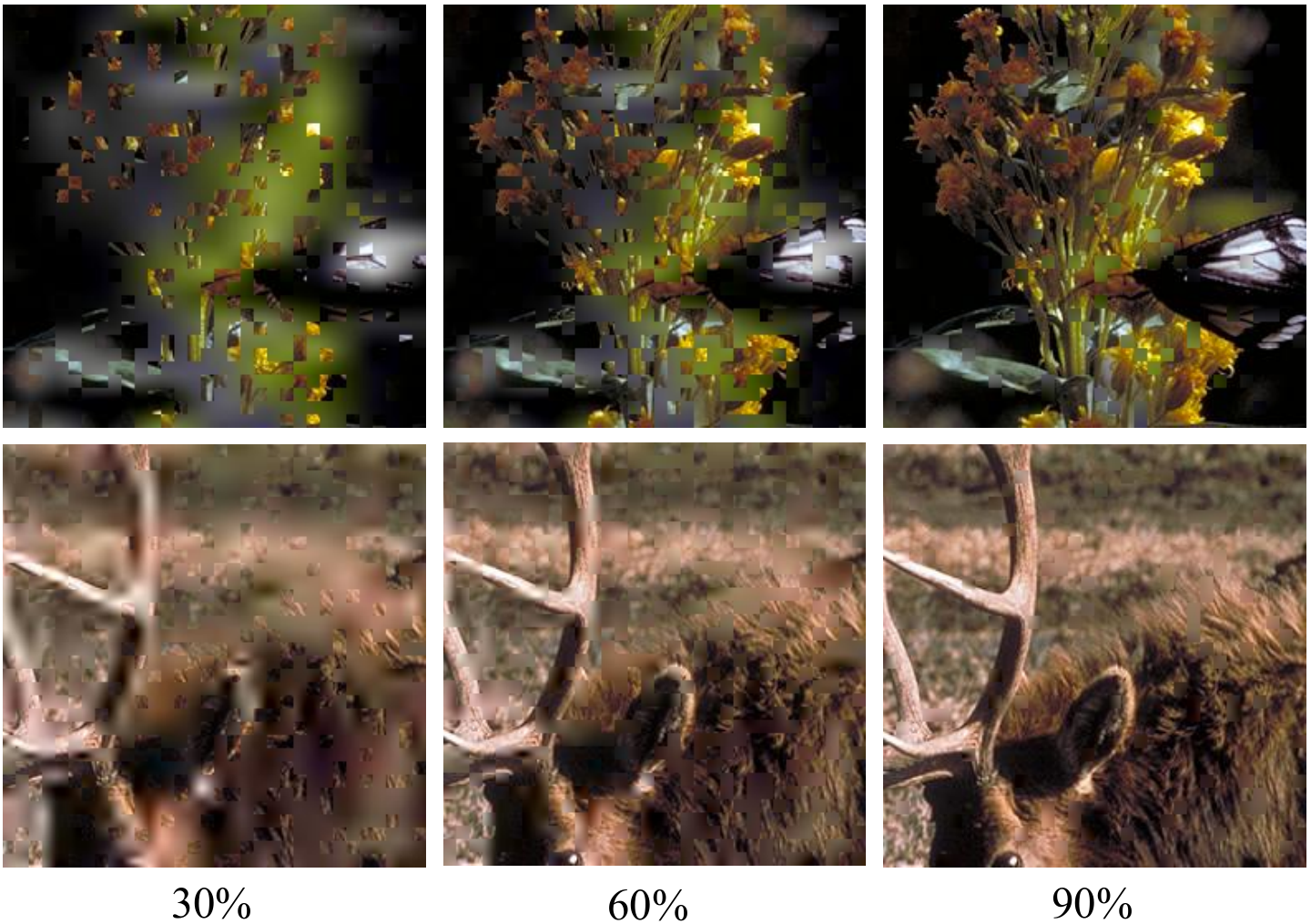}
    \caption{Sample display of masked images generated using different mask ratios. As the ratio of masked patches increases, more and more information is being recovered from the distorted image.}
    \label{fig:masksample}
\end{figure}

As shown in Table \ref{table_performance}, some representative IQA methods are selected for performance comparison, 
except for some methods to train models with external datasets to ensure fairness. 
The methods for comparison including hand-crafted based approaches, deep learning based NR-IQA approaches and deep learning based FR-IQA approaches.
Our methods are compared with these competitors on the four traditional IQA datasets, including LIVE \cite{sheikh2003image}, CSIQ \cite{larson2010most}, TID2013 \cite{ponomarenko2015image} and KADID-10K \cite{lin2019kadid}. We can observe that the FR-IQA models achieve a higher performance compared to the NR-IQA models, since the pristine-quality reference image provides more accurate reference information for quality assessment.

On LIVE, CSIQ and KADID-10k datasets, our MR-IQA achieves SOTA performance on all metrics.
Although WaDIQaM-FR achieves slightly better performance with our method on the TID2013 dataset, it is inferior to ours on the large KADID-10K dataset, indicating its limited generalization ability. 
To verify the potential of our method for quality assessment using masked images, we directly use random selection (Random) of positions instead of selecting mask positions based on difference (Diff) to generate masked images with the same proportion of reference images patches, although the performance is slightly worse than The latter, but still has significant advantages over other methods.
By adopting Mask strategy, our MR-IQA method achieves the best performance on all the four datasets. 
Especially on the larger KADID-10k database, we observe a solid improvement over previous work. 
Besides, even conventional IQA perform well on the smaller LIVE and CSIQ, but fall short on the more complex datasets such as TID2013 and KADID-10k. 
Furthermore, MR-IQA does not require input of complete reference image and is directly applicable to the simple FR IQA problem involving a pair of images. 
As such, we achieve both excellent performance and reduced input requirements for the FR-IQA method.

\begin{figure}[t]
  \centering
    \includegraphics[width=1.0\linewidth]{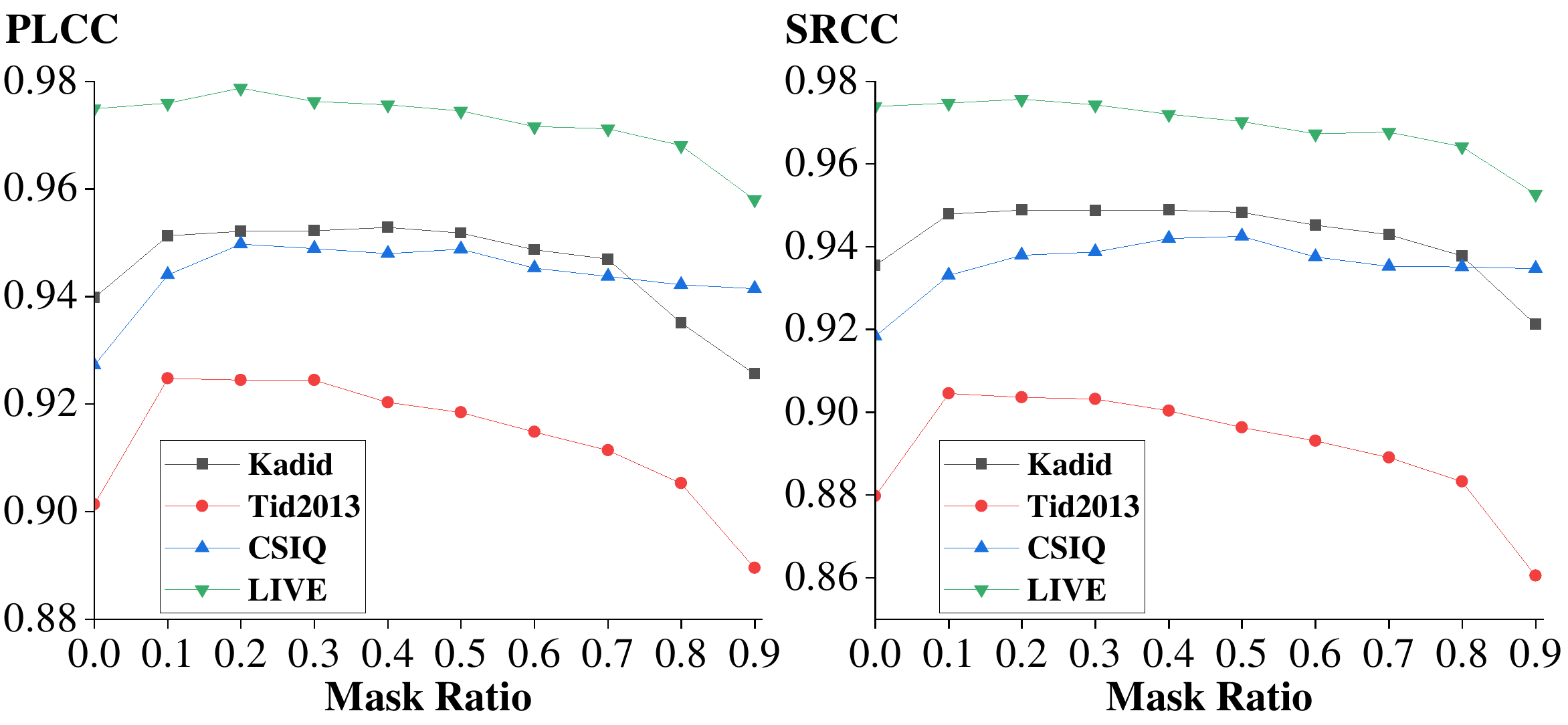}
    \caption{PLCC/SRCC results of different mask ratios on four IQA datasets. By introducing an appropriate ratio of masked patches, performance is significantly improved.}
    \label{fig:maskratio}
\end{figure}
\subsection{Patch Masking Strategies}
In order to explore the influence of the proportion of mask in the masked image on the evaluation results, 
we generate masked images with different proportions of reference image patches and test the effect. 
It should be noted that, in order to flexibly adjust the mask ratio, 
we do not generate a mask image based on the difference between the reference image and the distorted image, but randomly select locations to mask the distorted image using the reference image patches.
As shown in Fig. \ref{fig:masksample}, two images suffering from severe JPEG2000 compression distortion are masked by the reference image patches with different proportions (30\%, 60\%, 90\%).

We keep the rest of the model unchanged and verify the performance of the model on masked images with different mask ratios on four datasets. The SRCC and PLCC results are shown in Fig. \ref{fig:maskratio}.
Except that the performance of different mask ratios does not change significantly on the LIVE dataset with simpler distortion types, 
the addition of a small number of reference image patches improves the performance of the model, but the performance of the model decreases significantly when the mask ratio is too large.


\subsection{Cross-Database Performance Evaluation}
To verify the generalisation ability of our proposed MR-IQA method, the cross dataset test is performed to compare with BRISQUE 
 \cite{mittal2012no}, FRIQUEE \cite{ghadiyaram2017perceptual}, M3 \cite{xue2014blind}, CORNIA \cite{ye2012unsupervised}, HOSA \cite{xu2016blind}.
The cross database experiments are conducted by training the model on an entire dataset and testing it on the other two datasets.
Since scores from different databases have different scales and meanings, we add a linear operation to the subjective scores to evaluate SRCC.
We show the SRCC results in Table \ref{table_cross}, where the best results are shown in bold. 
It can be observed that the results of our proposed method significantly outperform other methods on all database, especially in a case that training on a small database with limited distortion types (LIVE, CSIQ) while testing on the TID2013.

\subsection{Ablation Study}
\begin{table}[]
\footnotesize
\begin{center}
\setlength{\tabcolsep}{1.3mm}{

\begin{tabular}{l|cc|cc|cc}
\Xhline{1.2pt} \hline \specialrule{0em}{1pt}{1pt}

\multicolumn{1}{c|}{Trained on:} & \multicolumn{2}{c|}{CSIQ}       & \multicolumn{2}{c|}{LIVE}       & \multicolumn{2}{c}{TID2013}     \\ \hline
\specialrule{0em}{1pt}{1pt}
\multicolumn{1}{c|}{Tested on:}  & LIVE           & TID2013        & CSIQ           & TID2013        & LIVE           & CSIQ           \\ \hline
\specialrule{0em}{1pt}{1pt}
\rowcolor[HTML]{EFEFEF} 
BRISQUE\cite{mittal2012no}                          & 0.847          & 0.454          & 0.562          & 0.358          & 0.790          & 0.590          \\
\specialrule{0em}{1pt}{1pt}
FRIQUEE\cite{ghadiyaram2017perceptual}                           & 0.879          & 0.463          & 0.722          & 0.461          & 0.755          & 0.635          \\
\specialrule{0em}{1pt}{1pt}
\rowcolor[HTML]{EFEFEF} 
M3\cite{xue2014blind}                               & 0.797          & 0.328          & 0.621          & 0.344          & 0.873          & 0.605          \\
\specialrule{0em}{1pt}{1pt}
CORNIA\cite{ye2012unsupervised}                           & 0.853          & 0.312          & 0.649          & 0.360          & 0.846          & 0.672          \\
\specialrule{0em}{1pt}{1pt}
\rowcolor[HTML]{EFEFEF} 
HOSA\cite{xu2016blind}                             & 0.594          & 0.361          & 0.594          & 0.361          & 0.846          & 0.612          \\ \hline
\specialrule{0em}{1pt}{1pt}
Ours                             & \textbf{0.882} & \textbf{0.592} & \textbf{0.757} & \textbf{0.618} & \textbf{0.926} & \textbf{0.910} \\ \Xhline{1.2pt} \hline
\specialrule{0em}{1pt}{1pt}

\end{tabular}
}
\caption{SRCC performance with cross-database performance evaluation on the  CSIQ \cite{larson2010most}, LIVE \cite{sheikh2003image} and TID2013 \cite{ponomarenko2015image} datasets.}
\label{table_cross}
\end{center}
\end{table}



We conduct the ablation experiment to verify the effectiveness of AGCS, MG, and FMM in our methods. 
Considering that the TID2013 \cite{ponomarenko2015image} and KADID-10k \cite{lin2019kadid} have more distorted images and the distortion types are more complex, all the ablation study experiments are performed on them.
We removed AGCS, MG, and FMM, only use the backbone to extract multi-layer features and then used the scoring module to complete the scoring as a baseline.
The performance contributions of each individual component are shown in Table \ref{table_ablation}.

To verify the performance of MG module which is the core design of our method, we remove this module, since the FMM module depends on the addition of reference image patches, it is also removed, only the AGCS module is used. 
Compared to our complete method, after adding the AGCS module to the baseline, the performance improves slightly, while removing the MG and FMM results in the greatest performance degradation on the two datasets.

Then, we removed the FMM and the AGCS module respectively on the basis of our complete method, SRCC and PLCC both drop slightly on TID2013 and KADID-10K.
Finally, when the three modules designed in this paper are added to the baseline, it achieves better performance than other ablation methods.

\begin{table}[]
\footnotesize
\begin{center}
\begin{tabular}{ccccc}
\Xhline{1.2pt} \hline \specialrule{0em}{1pt}{1pt}
                       &                      &                         & TID2013     & KADID-10K       \\ \specialrule{0em}{1pt}{1pt}
\multirow{-2}{*}{AGCS} & \multirow{-2}{*}{MG} & \multirow{-2}{*}{FMM} & SRCC / PLCC   & SRCC / PLCC   \\ \hline \specialrule{0em}{1pt}{1pt}
\rowcolor[HTML]{EFEFEF} 
\ding{55}                      & \ding{55}                    & \ding{55}                       & 0.873 / 0.895 & 0.932 / 0.936 \\ \specialrule{0em}{1pt}{1pt}
\ding{51}                      & \ding{55}                    & \ding{55}                       & 0.880 / 0.901 & 0.935 / 0.940 \\ \specialrule{0em}{1pt}{1pt}
\rowcolor[HTML]{EFEFEF} 
\ding{51}                      & \ding{51}                    & \ding{55}                       & 0.903 / 0.927 & 0.945 / 0.947 \\ \specialrule{0em}{1pt}{1pt}
\ding{55}                      & \ding{51}                    & \ding{51}                       & 0.909 / 0.925 & 0.947 / 0.952 \\ \specialrule{0em}{1pt}{1pt}
\rowcolor[HTML]{EFEFEF} 
\ding{51}                      & \ding{51}                    & \ding{51}                       & \textbf{0.912} / \textbf{0.930} & \textbf{0.952} / \textbf{0.955} \\ \Xhline{1.2pt} \hline \specialrule{0em}{1pt}{1pt}
\end{tabular}
\caption{SRCC/PLCC performance with ablation studies on the  TID2013\cite{ponomarenko2015image} and KADID-10K\cite{lin2019kadid} datasets.}
\label{table_ablation}
\end{center}
\end{table}


\section{Conclusion}
In this paper, we propose a Mask Reference IQA (MR-IQA) method to address the semantic and texture information loss in distorted images. To achieve this goal, we introduce a Mask Generator (MG) to mask specific patches of a distorted image and supplement missing patches with the reference image patches, then we use Mask Reference Network (MRNet) to shield the perturbation of masked patches and complete the quality assessment of the masked image. Compared with the FR-IQA and NR-IQA methods, our method achieves top performance on four representative datasets, and reduces almost 50\% of the computational complexity in structural design compared to learning-based FR-IQA methods. We hope that the MR-IQA will motivate the community to rethink IQA and stimulate research with a broader perspective.
\newpage



{\small
\bibliographystyle{ieee_fullname}
\normalem
\bibliography{egbib}
}

\end{document}